\documentclass[runningheads,a4paper]{llncs}
\usepackage{amssymb}
\usepackage{graphicx}
\usepackage{url}
\urldef{\mailsa}\path|{long.nguyen, dylan.salopek, zhou.yang,|
\urldef{\mailsb}\path|joshua.stuve, fang.jin}@ttu.edu|    

\let\OLDthebibliography\thebibliography
\renewcommand\thebibliography[1]{
  \OLDthebibliography{#1}
  \setlength{\parskip}{0pt}
  \setlength{\itemsep}{1pt plus 0.8ex}
}
\usepackage{enumitem}
\usepackage{graphicx}
\usepackage[usenames, dvipsnames]{color}
\usepackage{bm}

\usepackage{cite}
\usepackage{graphicx}
\usepackage{amsmath}
\usepackage{enumitem}
\setlist[enumerate]{itemsep=0mm}

\usepackage[norelsize]{algorithm2e}
\usepackage{color}
\usepackage{subfigure}
\graphicspath{{figures/}{pictures/}{images/}{./}} 
\usepackage{booktabs} 
\usepackage{multirow}
\usepackage{array}

\begin{document}

\mainmatter  

\title{Spatial-temporal Multi-Task Learning \\ for Within-field Cotton Yield Prediction}


%
%




%
%


\author{Long Nguyen\inst{1} \and
Jia Zhen\inst{2} \and
Zhe Lin \inst{1} \and
Hanxiang Du \inst{1} \and
Zhou Yang \inst{1} \and
Wenxuan Guo \inst{1} \and
Fang Jin\inst{1}
}

%
%
\institute{
Texas Tech University, Lubbock, Texas, USA 
\email{\{long.nguyen,zhe.lin,hanxiang.du,zhou.yang, wenxuan.guo, fang.jin\}@ttu.edu}\\
\and
George Washington University, Washington, USA\\
\email{jiazhen\_zhu@gwmail.gwu.edu}
}

\maketitle

\begin{abstract}
Understanding and accurately predicting within-field spatial variability of crop yield play a key role in site-specific management of crop inputs such as irrigation water and fertilizer for optimized crop production. However, such a task is challenged by the complex interaction between crop growth and environmental and managerial factors, such as climate, soil conditions, tillage, and irrigation. In this paper, we present a novel Spatial-temporal Multi-Task Learning algorithms for within-field crop yield prediction in west Texas from 2001 to 2003. This algorithm integrates multiple heterogeneous data sources to learn different features simultaneously, and to aggregate spatial-temporal features by introducing a weighted regularizer to the loss functions. Our comprehensive experimental results consistently outperform the results of other conventional methods, and suggest a promising approach, which improves the landscape of crop prediction research fields.     


\end{abstract}

\section{Introduction}\label{sec:Introduction}


Cotton is an important cash crop native to tropical and subtropical regions in the world. 
Accurate yield prediction not only provides valuable information to cotton producers for effective management of the crop for optimized production, but also is important to policymakers, as well as consumers of agricultural products.
However, cotton yield prediction is challenging due to complex interactions between crop growth and weather factors, soil conditions, as well as management factors, such as irrigation, tillage, rotation, etc. Moreover, simply applying other crop yield prediction models on cotton may lead nothing but disappointment: A prediction model that works on other crops like wheat, rice, and sugarcane, however, fails on predicting cotton yield\cite{bastiaanssen2003new}.   

\begin{figure}[t]
    \centering
    \includegraphics[width=240pt, height=140pt]{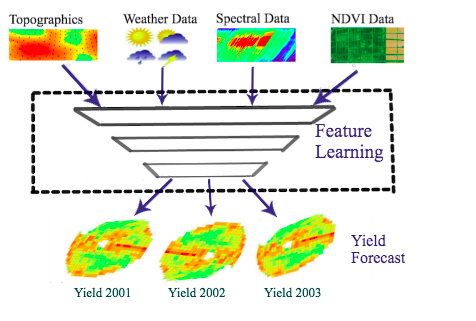}
    \caption{Data sources in our prediction model.}
    \label{fig:cotton-producing}
\end{figure}

The existing approaches estimate crop yield based either on the crop sown areas, crop-cutting experiments or market arrivals show wide variability because of their inability to capture the indeterminate nature of the crop and its response to environmental conditions~\cite{hebbar2008predicting}. One of the attempts is to apply Grey model~\cite{hsu2003applying, akay2007grey} on production prediction, utilizing short-term forecasting with exponential growth. The regression-based method such as time series analysis can also be applied to production prediction~\cite{battese1988prediction,sugihara1990nonlinear}, but it suffers from great variation when the external environment is under significant variation. Another frequently used method for prediction is the differential equation model~\cite{garcia1983stochastic}, which demands the system to be stable and requires extra work to solve the equation. 

Conventional yield prediction treats the field uniformly despite inherent variability. Uniform management may result in over- or under-application of resources in specific locations within a field, which may have a negative impact on the environment and profitability~\cite{guo2015data}. However, consistent and accurate within-field yield prediction is challenging due to the high accuracy requirement under the complex interactions between yield-influencing factors, such as soil, weather, water, and spatial correlations.

With the introduction of the global positioning system (GPS), geographic information systems (GIS), and yield monitors along with other new technologies, we can quantify spatial variability in soil properties and crop yield in small areas of a field. 
As satellite and drone technologies develop, we are able to collect remote sensing images at fine resolutions to support within-field yield forecast. Within-field scale crop yield prediction provides valuable information for producers to site-specifically manages their crop, which can optimize crop production for maximum profitability. In the within-field prediction procedure, we use a 30-m grid to represent a continuous surface. 

The advancement of machine learning offers a different approach compared with the traditional ways for yield forecasting. The rapid advances in sensing technologies, the use of fully automated data recording, unmanned systems, remote sensing through satellites or aircraft, and real-time non-invasive computer vision, are additional boosts for enabling the new yield forecasting model. Due to the capability of machine learning based systems to process a large number of inputs and handle non-linear tasks, people have attempted to predict county-level soybean yield in the United States \cite{jiang2018predicting,johnson2014assessment}. 
However, using deep learning for within-field cotton forecast remains as an untouched ground. 
In our work, the within-field forecasting is based on each individual grid for one field in West Texas area across three years (2001, 2002, 2003) in order to predict the cotton yield before harvest.

On this account, we propose a Multi-Task learning model to predict within-field cotton yield. 
As shown in Figure~\ref{fig:cotton-producing}, this model ingests many sources of data which contain features for different learning tasks, including soil topographic attributes (elevation, slope, curvature, etc.), spectral data (Blue, Green, Red, and NIR bands denoted as BAND1, BAND2, BAND3, and BAND4, respectively), normalized difference vegetation index (NVDI) during the crop seasons; and weather (temperature, rainfall, etc.) data. 
These multiple data sources are aggregated in the shared layer before transferring to task-specific layers. This type of design in a Multi-Task learning model makes it capable of enhancing specific learning task by utilizing all sources of information of other related tasks. In other words, this allows us to take various factors and variables into consideration to achieve a more accurate yield prediction. 
On the other hand, crop yield within a field  is typically autocorrelated, meaning yield values close together are likely more similar than those farther apart. Hence, to incorporate the spatial relationship, we propose a spatial regularization term to minimize the yield difference between one region and the weighted average of neighboring regions.

The main contributions of this paper are summarized as below:
\begin{itemize}
   \item   We design an innovative multi-task learning approach to predict within-field cotton yield across several years. Different from other machine learning models, to predict the cotton yield for a specific year is one of the tasks in our model; each individual task is enhanced through its access to all available data from prior years.
   \item   This work provides an entirely new vision for grid-scale crop yield prediction. To the best of our knowledge, this is among the first attempts to predict fine-grain cotton yield with the Multi-Task Learning approach, as existing work focus more on county-level or country level. 
   \item   We introduce a spatial weight regularizer in order to overcome the effects of geographical distance on yield prediction. Each grid is trained to minimize not only the difference between the prediction and the actual value, but also the difference between its yield and its neighbors.
   \item   We perform a comprehensive set of experiments using the real-world dataset that produced results consistently outperformed other competitive methods, which could provide guidance for achieving higher crop production.
\end{itemize}

\section{Related Work}
\vspace{-2mm}
\paragraph{\textbf{Crop Yield Prediction.}}
Crop yield prediction is challenging. Many researches are conducted based on NDVI derived from the new moderate resolution imaging spectroradiometer (MODIS) sensor \cite{mkhabela2011crop}, MODIS two-band Enhanced Vegetation Index \cite{bolton2013forecasting}, even future weather variables \cite{schlenker2009nonlinear}. Various methodologies are employed, such as statistical models \cite{lobell2010use}, fuzzy systems and Artificial Neural Networks \cite{dahikar2014agricultural}, deep long short-term memory model \cite{jiang2018predicting} and deep neural network \cite{oliveira2018scalable}. Ji et al. \cite{ji2007artificial} investigated the effectiveness of machine learning methods. Unfortunately, most of the academic endeavors centered on Artificial Neural Networks with one or a few data sources undermine their predict performance; Most previous studies assuming the crop yield uniformly distributed over space ignore the spatial variations. 

\vspace{-2mm}
\paragraph{\textbf{Multi-Task Learning.}}
Multi-Task learning (MTL) is implemented to predict spatial events due to its competence to exploit dynamic features and scalability \cite{zhao2015multi}. Through learning multiple related tasks simultaneously and treating prediction at each time point as a single task, MTL captures the progression trend of Alzheimer's Disease better \cite{zhou2011multi}. MTL also has outstanding performance in event forecasting across cities \cite{zhao2015multi} and fine-grain sentiment analysis  \cite{balikas2017multitask}, as well as in distance speech recognition \cite{zhang2017attention}. Lu et al. proposed a principled approach for designing compact MTL architecture by starting with a thin network and dynamically widening it in a greedy manner \cite{lu2017fully}. 
To the best of our knowledge, usage of MTL in crop yield forecast in within-field practice is untouched. 
We propose a Multi-Task Learning model which targets at predicting each grid in the field for a crop season as an individual task. Meanwhile, we incorporate the spatial correlations as a regularization term to minimize the prediction errors.

\section{Proposed Model}
\vspace{-2mm}
\subsection{Overview}
\begin{figure*}[t]
    \centering
    \includegraphics[width=\linewidth, height=125pt]{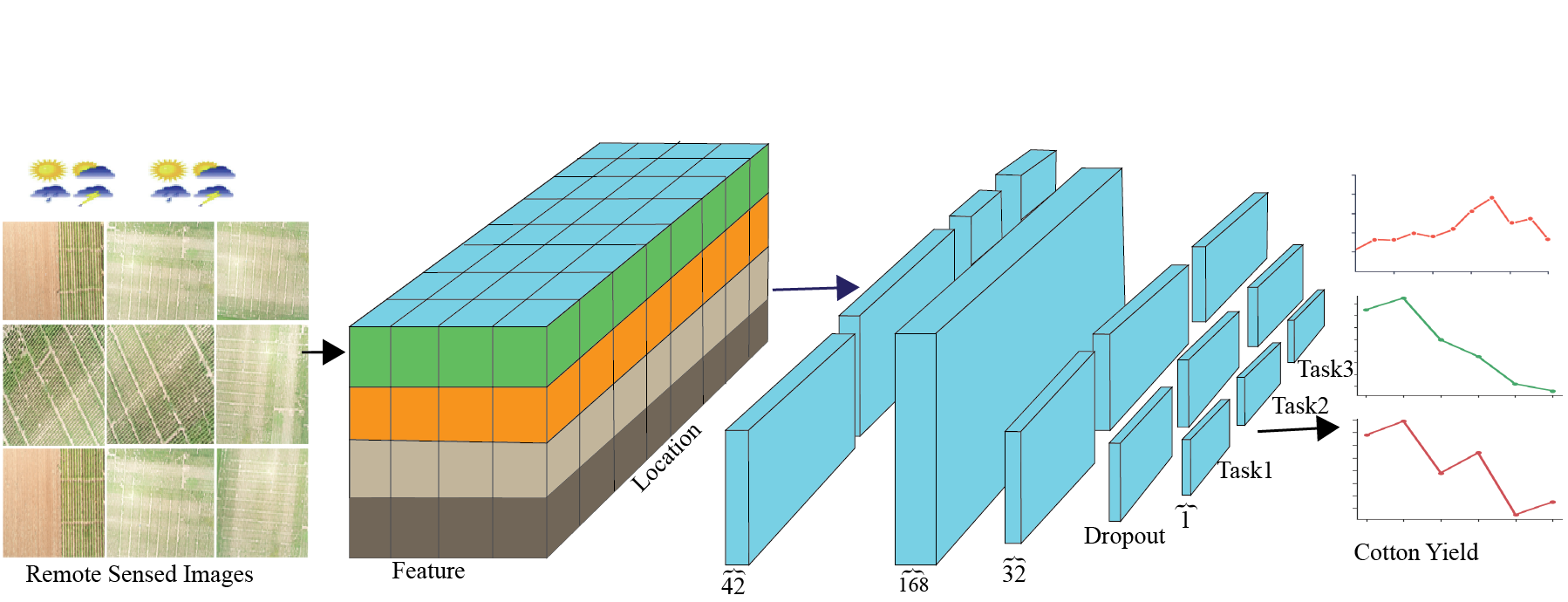}
    \caption{The cotton yield prediction framework}
    \label{fig:architecture}
\end{figure*}

Figure \ref{fig:architecture} presents the framework of our prediction model. In our model, cotton yield prediction of the year 2001, 2002 and 2003 are achieved in parallel. We utilized the Dense and Dropout layers in the network. 
A shared Dense layer is used to extract latent features from all data dimensions, which are aggregated and fed into multiple sub-networks for yearly yield forecasting task.
This aggregation is shared among all task-specific sub-networks. Therefore, it helps the task-specific sub-network to learn features from other tasks and to enhance its own prediction performance. Dense layer helps receive input from all the neurons in the previous layer with the intuition that all factors contribute each layer output neurons. Mathematically, the latent feature $\hat{p}^t$ learned after a fully connected layer is computed as: 
\begin{equation}
    \hat{p}^t = \sigma(\sum^{N}_{i=1}{x_i*w_i} + b )
\end{equation}
\vspace{-1mm},
 where $x_i$ represents input feature, $w_i$ is a weight element,  $b$ is a bias and $\sigma$ is the activation function. In our setting, $\sigma(x)$ is a Sigmoid function defined as $\sigma(x) = 1 / (1 + e^{-x})$. Moreover, the dropout layer or dropout regularization is also used to randomly exclude some neurons (in our setting, we exclude $20\%$ of the neurons) to avoid over-fitting. 


\vspace{-2mm}
\subsection{Cotton Yield Prediction}
Our feature set is enriched by concatenating the latent features and feeding the output into a shared Dense layer. Suppose $p^{t_1}, p^{t_2}, p^{t_3}$ and $p^{t_4}$ are features from our sources, the joint feature $v_{fc}$ is the concatenation (denote as $\oplus$) of those features~\cite{yao2018deep}: 
\begin{equation}
    v^{fc} = p^{t_1} \oplus p^{t_2} \oplus p^{t_3} \oplus p^{t_4}.
\end{equation}
Stacked on the top of the shared Dense layer are three separate sub-networks, and each is used for one yearly cotton yield forecasting task, as shown in Figure \ref{fig:architecture}. After this layer, the latent feature is learned at time $j$ following the equation: $h_j = \sigma(W_j*v^{fc}_j + b)$. 
Because cotton is usually planted by the end of May and harvest at the end of September or early October, we cut cotton's life cycle into several pieces, and each piece represents $2$ weeks. 
Instead of taking it as time series data, we treat it as a couple of separate temporal features and utilize fully connected Dense layers and Dropout layers behind the shared Dense layer. We define the regression function for task $t$ as:
\begin{equation}
\hat{y}^{t}_j = \sigma(W^{t}_{j}*h^{t}_{j} + b^{t}_j)
\end{equation}, where $W^{t}_{j}$ and $b^{t}_{j}$ are learnable parameters, $h^{t}_{j}$ is the output of the last hidden layer, and $\sigma$ is a linear activation function. The model output lies in the interval $[0, 1]$ after value normalization. We will recover them to the original values when doing performance evaluation. 

\vspace{-2mm}
\subsection{Spatial Feature in the Loss Function}
A loss function is defined as the mean square error between the observation and the prediction: 
\begin{equation}
    \mathcal{L(\theta)} = \sum_{k=1}^{N}{(y_k - \hat{y}_k)^2} 
\end{equation}, where $\theta$ means all learnable parameters, $N$ is the number of regions in the field, $y_k$ represents actual yield value, and $\hat{y}_k$ represents the predicted yield value. 
To train $\theta$ by minimizing the loss function may introduce overfitting. 
Therefore, for the grid-scale crop forecasting within a field, the spatial correlations depend heavily on the factor of distance.
This drives us to add a regularization term that the yield difference between the predicting region and the weighted average yield of the neighboring regions should be minimized. 
In particular, suppose $G(k)$ is the set of neighbors of region $k$, and $w(k,j)$ is the inverse distance weight between region $k$ and region $j$, the loss function now becomes:

\begin{equation}
    \mathcal{L(\theta)} = \sum_{k=1}^{N}{[(y_k - \hat{y}_k)^2} +  \lambda\sum_{j \in G(k)}{ \frac{w(k,j)}{|G(k)|}*|\hat{y}_k - y_j|^2]}
\end{equation}, where $\lambda$ is the hyper parameter, $d(k,j)$ is the Euclidean distance between these regions $k \neq j$. The spatial weight $w(k,j)$ is computed as:
\begin{equation}
    w(k,j) = \frac{1}{d(k,j)^p}
\end{equation}, where $p$ is the power parameter (which equals to $2$ in our experiment).

\vspace{-2mm}
\section{Experiments}
\vspace{-2mm}
\subsection{Dataset and Feature Extraction}
\begin{figure}[t]
    \centering
    \includegraphics[width=\linewidth,height=1.5in]{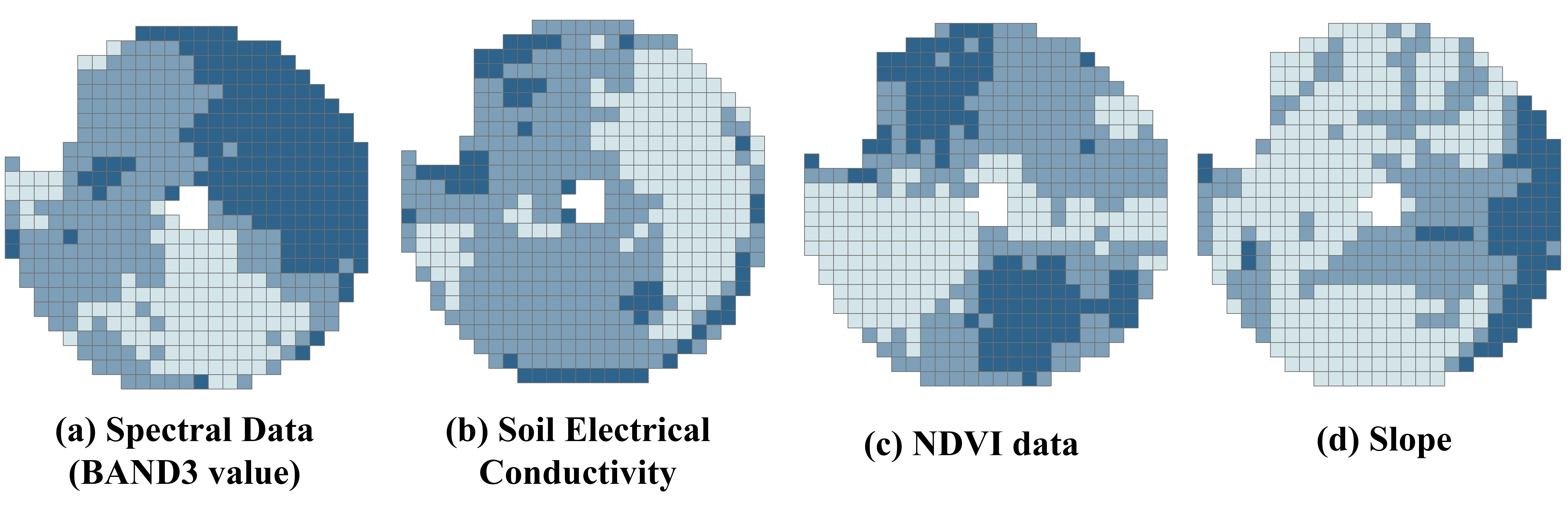}
    \caption{(a) to (d) Within-field feature value distributions. Darker colors indicate higher values. Each feature value is normalized into [0,1].}
    \label{fig:feature_heatmap_Hor1}
\end{figure}

Our dataset includes weather data, soil properties, spectral data, and NDVI. Spectral data and NDVI are extracted from Lantsat 5 and Landsat 7 remote sensing images.  
The multi-spectral images were collected from 2001 to 2003 of a cotton field in west Texas. The total area is approximately 48 ha. The sensed images spatial resolution is 30 m. Hence, there are 475 grid cells under investigation. Figure \ref{fig:feature_heatmap_Hor1} shows the distribution of some features over the field. 
  

\vspace{-2mm}
\paragraph{\textbf{Weather data.}} Weather data includes the daily temperature and rainfall level. For simplicity, we use the average of every two weeks' weather data as features to match the sensed images. 
\vspace{-2mm}
\paragraph{\textbf{Soil properties.}}
Topographic variation is a common characteristic of large agricultural fields that has effects on spatial variability of soil water and ultimately on crop yield \cite{hanna1982soil}. 
Besides, soil electrical conductivity (ECa) is also a reliable measurement of field variability.
The relationship between ECa and crop yield depends on climate, crop type and other specific field conditions \cite{kitchen1999soil}. The variables that are considered in this paper include elevation, slope, curvature, the average electrical conductivity of soil, etc.



\vspace{-2mm}
\paragraph{\textbf{Field spectral data before planting.}}
Field spectrum before planting may influence the entire crop yield. This data is composed of four spectral bands extracted from the sensed images with 30-metre spatial resolution. Band1, Band2, Band3, and Band4 represent blue, green, red and near infrared value, respectively.  
\vspace{-2mm}
\paragraph{\textbf{NDVI data.}}
NDVI represents Normalized Difference Vegetation Index. It \textit{is typically related to amount or density of vegetation, which is calculated as the difference between the reflectance in near-infrared (which vegetation strongly reflects) and red wavelengths divided by the sum of these two}. 
 NDVI is computed as: $NDVI = \frac{NIR-RED}{NIR+RED}$,
where $NIR$ represents the spectral reflectance in near-infrared wavelength and $RED$ is the spectral reflectance in the red wavelength. 

\vspace{-2mm}
\subsection{Competing Approaches and Comparison Metrics}
\subsubsection{Competing Approaches:}
In our experiment, a list of classical forecasting models are used for comparison and analysis:
\vspace{-2mm}
\paragraph{\textbf{Linear Regression}}
This is a traditional linear regression model. Its standard formula is: $ y = \sum_{i=1}^{n} \alpha_i x_i   + \epsilon$
where $y$ is the response variable, $x_i$ is the feature and $\epsilon$ is the deviation.
\vspace{-2mm}
\paragraph{\textbf{Decision Tree Regression}} This method builds regression models in the form of a tree structure. It breaks down a dataset into smaller and smaller subsets while an associated regression tree is incrementally developed at the same time. The splitting point is chosen with the smallest sum of squared error (SSE) between predicted values and actual values. In the case of a regression model, the target variable does not have class.
\begin{figure}[t]
    \centering
    \includegraphics[width=3in, height=1.6in]{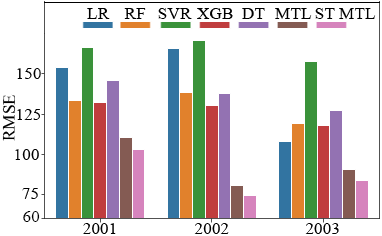}
    \caption{Performance comparison in each year measured in RMSE, using whole dataset until September.}
    \label{fig:prediction-year}
\end{figure}

\begin{figure}[t]
    \centering
    \includegraphics[scale=0.6]{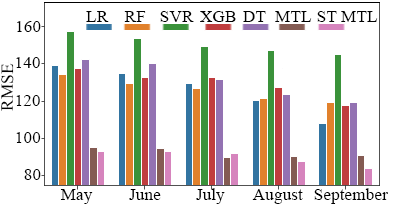}
    \caption{Model performance in each month of 2003 measured in RMSE.}
    \label{fig:prediction-month}
\end{figure}

\paragraph{\textbf{Random Forest}}
This model tries to fit a number of regression trees on various sub-samples of the dataset and uses averaging to improve the predictive accuracy and control overfitting. Each leaf of the tree contains a distribution for the continuous output variable.
\vspace{-2mm}
\paragraph{\textbf{Support Vector Regression (SVR)}}
 SVR is a nonparametric technique that 
 aims to find a function $f(x)$ that produces output deviated from observed response values $y_n$ by a value no greater than $\epsilon$ for each training point $x$, and meanwhile, as flat as possible.
 \vspace{-2mm}
\paragraph{\textbf{XGBoost}}
This is an extension of gradient boosting machine (GBM) algorithm that tries to divide the optimization problem into two parts by first determining the direction of the step and then optimizing the step length.

\vspace{-2mm}
\subsubsection{Evaluation Metrics}
Let $N$ be the number of grids under forecast. We denote $A_i$ as the actual crop yield and $F_i$ is the forecast yield for grid $i$. A set of classical metrics such as Mean square error (MSE), Root mean square error (RMSE), Mean absolute error (MAE), Mean absolute percentage error (MAPE) and Max error (ME) are used to elaborate the performance. These measures are computed as: $ MSE = \frac{1}{N} \sum_{i=1}^{N} (A_i - F_i)^2$, $RMSE = \sqrt{MSE}$, $MAE = \frac{1}{N} \sum_{i=1}^{N} |A_i - F_i|$, $MAPE = \frac{100\%}{N} \sum_{i=1}^{N} \frac{|A_i - F_i|}{A_i}$ and $ ME = max(|A_i - F_i|)$ where $i = 1,...,N$.


\begin{figure}[t]
    \centering
    \includegraphics[width=\linewidth,height=2.5in]{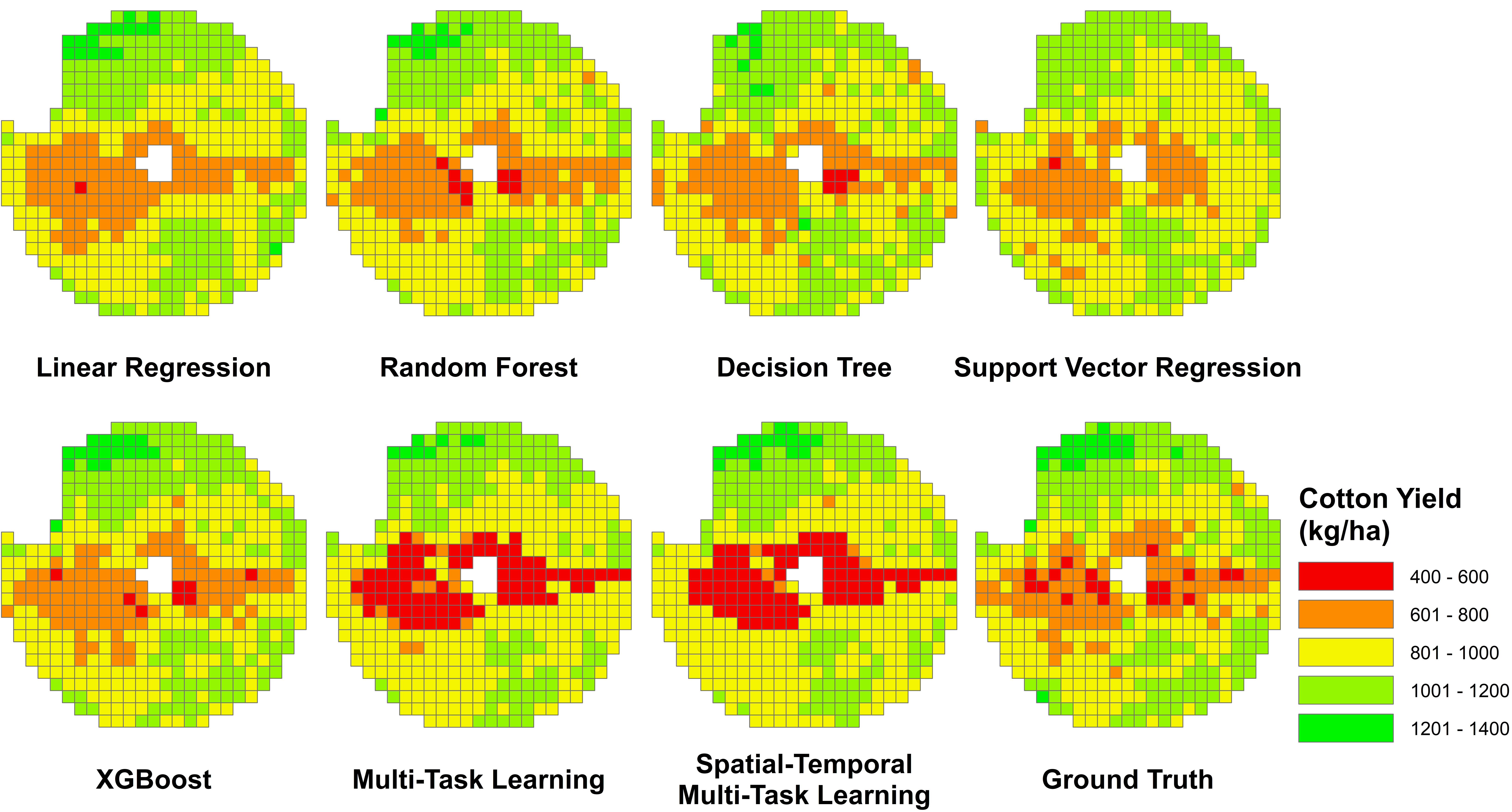}
    \caption{Within-field cotton yield prediction on different algorithms versus ground truth yield monitor data in 2003.}
    \label{fig:8_yields1}
\end{figure}

\begin{table*}[t]
\setlength{\tabcolsep}{5pt}
  \centering
 \caption{Cotton yield prediction performance comparison of year 2003 while combining all data sources. Bold values represent the best results.}    
 \begin{tabular}{lr | r | r | r | r}
    \toprule
          & \multicolumn{1}{ c | }{MSE} & \multicolumn{1}{| c |}{RMSE} & \multicolumn{1}{| c |}{MAE} & \multicolumn{1}{| c }{MAPE} & \multicolumn{1}{| c }{Max Error} \\
  
    \midrule
    Linear Regression & 17,875.6 & 133.7  & 109.0 & 10.96 & 276.7 \\
    \midrule
    Random Forest & 19,295.3 &  138.9 & 112.0 & 10.68 & 321.7 \\
    \midrule
     Decision Tree & 22,320.6 &  149.4 & 118.6 & 12.50 & 394.2 \\
    \midrule
    Support Vector Regression & 33,588.5 & 183.3  & 142.2 & 13.35 & 475.4 \\
    \midrule
    XGBoost & 19,498.4 &  139.6 & 111.4 & 10.05 & 353.0 \\
    \midrule
    Multi-Task Learning & 8,267.5 & 90.9  & 70.5 & 8.08 & 256.2 \\
    \midrule
    Spatial-Temporal M.T.L. & \textbf{7,013.5} & \textbf{83.7} & \textbf{63.6} & \textbf{7.55} & \textbf{254.4} \\
    \bottomrule
    \end{tabular}
  \label{table:performance-comparison}
\end{table*}

\subsection{Experimental Results}
\subsubsection{Average Performance}

\begin{table*}[t]
\setlength{\tabcolsep}{10pt}
  \centering
 \caption{Performance of the spatial-temporal Multi-task Learning model using individual source of data.}    
 \begin{tabular}{l| r | r | r | r | r}
    \toprule
          \multicolumn{1}{ c | }{Input Source}
          & \multicolumn{1}{ c | }{MSE} & \multicolumn{1}{| c |}{RMSE} & \multicolumn{1}{| c |}{MAE} & \multicolumn{1}{| c }{MAPE} & \multicolumn{1}{| c }{Max Error} \\
    \midrule
    Soil Properties & 15,791.9 & 125.7 & 98.2 & 12.04 & \textbf{372.7} \\
    \midrule
    Spectral Data & 21,292.5 & 145.9 & 114.4 & 13.62 & 450.1 \\
    \midrule
    NDVI & \textbf{15,134.5} & \textbf{123.0} & \textbf{94.7} & \textbf{11.80} & 470.4 \\
    \bottomrule
    \end{tabular}
  \label{table:performance-one-source-comparison}
\end{table*}

\begin{table*}[h]
\setlength{\tabcolsep}{9pt}
  \centering
 \caption{Discover the importance of sources of data by removing one source at a time.}    
 \begin{tabular}{l| r | r | r | r | r}
    \toprule
        \multicolumn{1}{ c | }{Removed Source}
          & \multicolumn{1}{ c | }{MSE} & \multicolumn{1}{| c |}{RMSE} & \multicolumn{1}{| c |}{MAE} & \multicolumn{1}{| c }{MAPE} & \multicolumn{1}{| c }{Max Error} \\
 
    \midrule
    Soil Properties & 13,383.0 & 115.7 & \textbf{88.8} & \textbf{10.79} & 481.6 \\
    \midrule
    Spectral Data & \textbf{13,189.6} & \textbf{114.8} & 93.0 & 11.54 & \textbf{276.5} \\
    \midrule
    NDVI & 15,642.0 & 125.1 & 100.3 & 12.39 & 330.0 \\
    \bottomrule
    \end{tabular}
  \label{table:performance-multi-sources-comparison}
\end{table*}

\begin{figure}[t]
    \centering
    \includegraphics[width=\linewidth,height=1in]{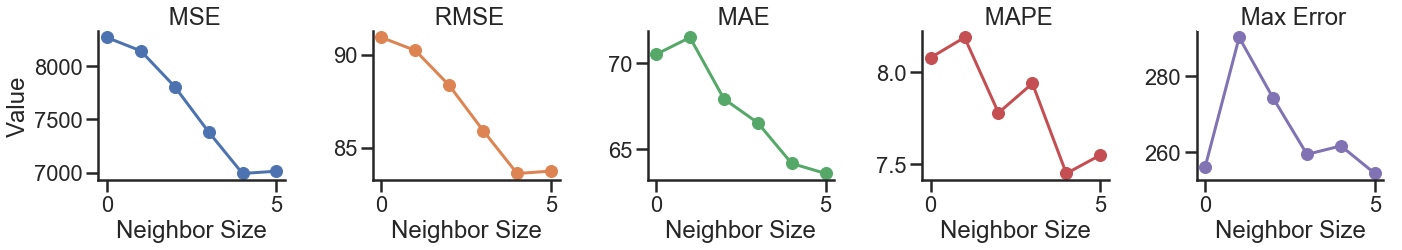}
    \caption{Impact of neighborhood size on model performance}
    \label{fig:impact-neighborhood-size}
\end{figure}
Figure \ref{fig:prediction-year} shows the performance of our proposed model compared with other baselines in term of RMSE metric. We can see that, the Multi-Task learning and our proposed Spatial-Temporal Multi-Task learning model have the least error in all three years.  
Figure \ref{fig:8_yields1} and Table \ref{table:performance-comparison} take yield prediction performance of 2003 as an example. Again, the Multi-Task Learning and our Spatial-Temporal Multi-Task learning methods show significant superiority than all the other approaches. With the whole data package, our model achieves the smallest error metrics (MSE, RMSE, MAE, MAPE, ME) which are $7,013.5$, $83.7$, $63.6$, $7.55$ and $254.4$, respectively. The overall performance of normal Multi-Task Learning is second and close to that of our proposed approach, while Support Vector Regression shows the worst performance. 

\vspace{-2mm}
\subsubsection{Real-time Prediction Throughout the Year}
Considering the life cycle of cotton in the U.S., we train the model using partially available input features and predict the cotton yield in each month in an online manner.
Figure~\ref{fig:prediction-month} shows the performance when we try to make a prediction in May, June, July, August and September, using only the data available up to that point.
As more information is available, most of the models improve. The improvement during the first three months is less than that of later two months. All models perform better in August and reach the best in September.

\vspace{-2mm}
\subsubsection{Spatial Correlations:}
We also vary the neighborhood distance of each region from $1$ to $5$ to verify the impact of spatial correlation among regions under prediction. As shown in Figure \ref{fig:impact-neighborhood-size}, MSE, RMSE and MAE gradually decrease when the distance increases. This trend stops when neighborhood size equals to $4$. The performance becomes more stable afterwards. Even though we see a random value in MAPE and ME metrics with respect to the neighborhood distance, there is also a decreasing trend on MAPE and ME when neighborhood size increases. Therefore, in our experiment, we set the neighbor distance as 5. 

\vspace{-3mm}
\subsubsection{Understanding the Importance of the Features:}
Since weather data is shared in all regions under prediction, we do not evaluate the impact of this feature in our model. Instead, we explore the impacts of soil properties, spectral conditions before planting and NDVI on the cotton yield prediction. We split the data by dimensions and conduct two experiments: the first experiment uses only one dimension feature at a time, the second experiment removes one dimension and use the rest input each time. 
The comparison and performance are listed in Table \ref{table:performance-one-source-comparison} and Table \ref{table:performance-multi-sources-comparison}, respectively.

From Table \ref{table:performance-one-source-comparison}, we see that using NDVI data helps to gain better precision compared with using the soil properties data alone and spectral data before planting alone in most of the metrics. It gets MSE and RMSE values at $15,134.5$ and $123.0$ while the spectral data produces the worst results, whose MSE and RMSE are $21,292.5$ and $145.9$, respectively. On the other hand, Table \ref{table:performance-multi-sources-comparison} indicates that if we ignore the spectral feature while keeping the rest, the model achieves the best results compared with ignoring the soil properties and NDVI features. These results demonstrate that the NDVI impacts the prediction the most, then soil properties, while spectral data before planting has the minimal impact. 

\vspace{-4mm}
\section{Conclusion}
\vspace{-2mm}
This paper proposes a novel Multi-task Learning framework for within-field scale cotton yield prediction, which ingests multiple heterogeneous data sources, such as soil type, weather, topographic, and remote sensing, and is capable of predicting within-field cotton yield throughout the growing season. By aggregating these multiple data sources in the shared layer before transferring to task-specific layers, this creative strategy is able to enhance specific learning task by utilizing sources from other related tasks. To minimize the spatial errors in prediction, this work introduces a spatial regularization to measure the correlations between a certain grid and its neighboring grids. The experimental results show the proposed approach consistently outperforms other competing approaches, and has a promising future in the crop yield prediction research field. 
\vspace{-2mm}
\bibliographystyle{IEEEtran}
\bibliography{reference}

\end{document}